\documentclass[10pt,twocolumn,letterpaper]{article}
\usepackage{booktabs}
\usepackage{multirow}
\usepackage[subtle]{savetrees}

\usepackage[pagenumbers]{wacv} 

\usepackage{graphicx}

\definecolor{wacvblue}{rgb}{0.21,0.49,0.74}
\usepackage[pagebackref,breaklinks,colorlinks,allcolors=wacvblue]{hyperref}
\usepackage{booktabs, multirow, amsmath, amssymb, url}

\title{AIRRI: Towards a Human-Readable, Machine-Resistant Reading Interface}

\author{
Emory Salaun 
\qquad
Syed Ayon 
\qquad
Kaleel Mahmood\thanks{Denotes Equal contribution.} 
\qquad
Shaun Wallace \footnotemark[1]
\\[0.5em]
Department of Computer Science and Statistics\\
University of Rhode Island\\
Kingston, Rhode Island, USA
\\[0.5em]
{\small
    Corresponding author: emory\_salaun@uri.edu
}
\\[0.5em]
}

\begin{document}
\maketitle
\begin{abstract}
The rapid advancement of image-to-text extraction systems and large language models (LLMs) poses new challenges for educational assessment by making assignment materials easy to transcribe and solve with minimal human effort. To address this problem, we present AIRRI (Artificial Intelligence Resilient Reading Interface), a framework for generating machine-resistant educational content, a dataset of higher-education assignments for evaluation, and an initial evaluation of the human readability of AIRRI machine-resistant text designs. AIRRI employs two categories of machine-resistance techniques. First, AIRRI applies image transformations including character spacing, word spacing, line spacing, character opacity variation, stripe overlays, and image noise. Second, AIRRI applies decision-based adversarial attacks: RayS, SurFree, and ADBA. We evaluate AIRRI against three image-to-text systems: Tesseract, EasyOCR, and the transformer-based GOT-OCR2 model. Our results show that AIRRI substantially degrades automated text extraction performance. AIRRI increased character error rates for EasyOCR (6.9\% to 55.9\%), Tesseract (2.3\% to 100.0\%), and GOT-OCR2 (30.7\% to 48.1\%). Word error rates increased for EasyOCR (24.9\% to 96.7\%) and Tesseract (10.7\% to 100.0\%). Among the evaluated decision-based attacks, ADBA consistently produced the largest degradation in image-to-text performance, while RayS and SurFree produced more moderate reductions. EasyOCR and Tesseract were generally more susceptible to both image-based and adversarial perturbations than the transformer-based GOT-OCR2 model. Our findings suggest that AIRRI can significantly reduce the reliability of image-to-text-to-LLM workflows, increasing the effort required for AI-assisted assignment completion. Anonymous code is available \href{https://anonymous.4open.science/r/AIRRI-C13C}{here}.

\end{abstract}

\section{Introduction}
Digital text is designed to maximize accessibility~\cite{text_accessibility} and machine readability~\cite{information_retrieval}, enabling copying, searching, indexing, and automated processing~\cite{beier2022readability}. While these properties provide significant usability benefits~\cite{wallace2022towards}, they also make content easily subject to automated extraction. Advances in image-to-text and language-processing systems have made it increasingly possible to rapidly convert on-screen text into machine-readable representations and process it automatically~\cite{afkari2025comprehensive,ocr,raisi2020text,wang2011end}. As artificial intelligence systems continue to improve, this raises an important question: can educational material remain readable and usable for human learners while becoming more resistant to automated extraction and machine-assisted completion?

Modern LLMs can generate human-like responses, summarize content, and answer questions \cite{gpt3, academic_integrity}. Combined with image-to-text extraction systems, they enable educational material to be rapidly converted into machine-readable text and processed through automated workflows. Thus, assignment content can be extracted from screenshots and given to AI systems for question answering, summarization, or assignment completion with minimal effort \cite{academic_integrity2}. This creates a growing challenge for educators seeking to promote direct student engagement while limiting low-effort AI-assisted workflows.

Existing responses to AI-assisted academic misuse frequently focus on behavioral policies\cite{genai_guidelines}, plagiarism detection, browser restrictions, or later identification of AI-generated work. While these approaches may discourage misuse or identify suspicious outputs after submission, they often do not address how easily students can extract and process educational content through image-to-text and LLM-based workflows. Furthermore, detection-based approaches may become less reliable\cite{detecting_ai} as generative models continue to improve and student usage patterns evolve. This motivates exploring preventive approaches that increase the difficulty of automated extraction while preserving the usability of instructional materials.

To address these challenges, we propose AIRRI (Artificial Intelligence Resilient Reading Interface). A system designed to increase resistance to automated extraction while preserving the readability of educational material for human users. AIRRI transforms assignment and instructional content into a machine-resistant format that impedes image-to-text recognition and downstream automated processing. This increases the difficulty of submitting extracted material directly to LLMs. AIRRI could enable educators to distribute content that remains available to human readers while making low-effort image-to-text and LLM-assisted workflows harder. Rather than eliminating misuse entirely, AIRRI increases the human effort required for such behavior. A student supplying assignment content to an LLM would need to manually transcribe it rather than rely on automated extraction methods. We investigate whether controlled text perturbations can reduce the effectiveness of image-to-text and downstream automated processing while using levels of perturbation to preserve some readability. We evaluate AIRRI across multiple image-to-text systems using real-world educational assignments and quantitative performance metrics to examine the trade-off between machine resistance and the severity of text perturbation. 
    
\begin{figure*}[t!]
\centering
\includegraphics[width=0.8\textwidth]{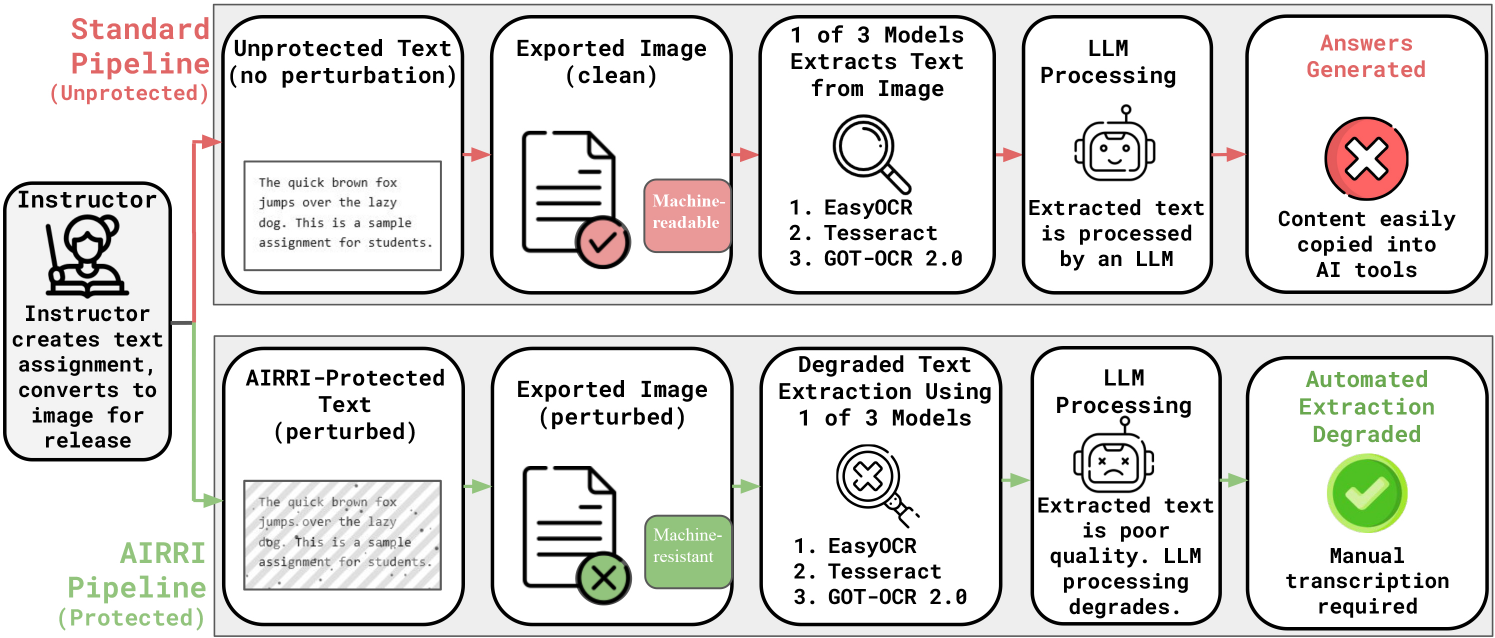}
\caption{The AIRRI pipeline transforms instructor-provided assignments into machine-resistant rendered text while preserving human readability. Image-to-text and LLM systems can extract unprotected text, whereas AIRRI-protected text requires manual transcription.}
\label{fig:airri_pipeline}
\end{figure*}
    
\section{AIRRI Pipeline}


AIRRI renders educational content as images and applies a combination of typographic, image-level, and adversarial perturbations. We cover AIRRI's threat model, pipeline architecture, rendering, defense mechanisms, and evaluation.

\subsection{Threat Model}

AIRRI assumes an adversary who attempts to process educational material through an image-to-text extraction pipeline followed by an LLM with minimal manual effort. The adversary is assumed to have access to modern automated text-extraction systems and may capture assignment content via screenshots or other extraction methods. Unlike traditional adversarial machine learning~\cite{traditional_adversarial}, AIRRI does not seek to induce specific misclassifications. Instead, it seeks to degrade automated text extraction while maintaining some usability for human readers.

An \textbf{attack} succeeds when an image-to-text system extracts enough information to reconstruct the original text with little or no manual transcription. A \textbf{defense} succeeds when it substantially degrades extraction quality while preserving human readability. AIRRI does not seek to make extraction impossible; instead, it increases the effort required for automated processing and reduces the effectiveness of AI-assisted extraction.

\subsection{Pipeline Overview}

Figure~\ref{fig:airri_pipeline} shows an overview of the AIRRI pipeline. Given an input assignment or instructional document, AIRRI produces a protected image intended to disrupt automated image-to-text extraction. AIRRI implements three categories of perturbations: typographic, image-level, and adversarial. The effectiveness of these perturbations is evaluated using EasyOCR\cite{easyocr}, Tesseract\cite{tesseract}, and GOT-OCR2\cite{gotocr}.

The AIRRI pipeline has five stages: \textbf{(1) Input Acquisition}, \textbf{(2) Text Rendering}, \textbf{(3) Perturbation Application}, \textbf{(4) Automated Extraction}, and \textbf{(5) Evaluation}. During Stage 1, instructor-provided educational material is the source text. Stage 2 renders the source material to a canvas-based image using fixed rendering parameters. Stage 3 applies the selected perturbation techniques to generate a protected image. Stage 4 evaluates extraction resistance by processing both protected and unprotected images through three image-to-text systems: EasyOCR, Tesseract, and GOT-OCR2, representing traditional and transformer-based extraction architectures. Stage 5 evaluates extraction quality and downstream assignment completion using Character Error Rate (CER), Word Error Rate (WER)~\cite{kiss2019ocr,2021ocrCER}, and LLM Error Rate (LER).

\subsection{Text Generation Pipeline}

AIRRI transforms instructor-provided educational material into a machine-resistant image representation using an HTML canvas rendering pipeline. Input text is rendered using the font Roboto at 16\,px with configurable character, word, and line spacing. We selected Roboto because it is the most viewed font on the web per Google Font Analytics~\cite{googlefontsanalytics,wallace2022towards} and is often the sole font studied in related readability research~\cite{palmen2023bold,llana2025typography,llana2026reading,nguyen2026rears}. Text is rendered with automatic line wrapping to preserve paragraph structure and reading order before exporting as a PNG image. For experimental consistency, all renderings were generated in a web browser at a fixed resolution of 1920$\times$1080. Clean and protected images were produced from identical source text and rendering settings, differing only through the application of AIRRI perturbations.

\subsection{Defense Mechanisms}

AIRRI employs three categories of defense mechanisms: typographic perturbations, image-level perturbations, and adversarial perturbations. Typographic perturbations modify the appearance and layout of rendered text, image-level perturbations alter the rendered image, and adversarial perturbations introduce model-targeted modifications intended to reduce extraction accuracy.

\subsubsection{Typographic Perturbations}

\begin{figure}[b!]
\centering
\includegraphics[width=0.9\linewidth]{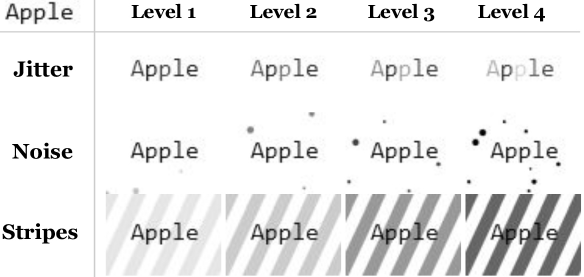}
\caption{Examples of AIRRI image-level perturbations; top left is the clean image. Intensity increases from Level 1 to Level 4.}
\label{fig:image_perturbation_levels}
\end{figure}

AIRRI applies four configurable typographic perturbation techniques to alter text appearance and reduce image-to-text extraction quality. Our text designs are based on prior readability literature to try and preserve human readability~\cite{wallace2022space,niklaus2023digital}. Prior research shows that typographic attributes such as font selection and character/word spacing can influence reading performance and readability \cite{cai2022personalized,suhail2026personalizing,suhail2026fontattributes}. Unless otherwise specified, each perturbation is applied uniformly throughout the rendered document. Figure~\ref{fig:image_perturbation_levels} shows examples of the image-level perturbations used in AIRRI at increasing intensity levels.

During rendering, \textbf{spacing (character, word, and line)} are CSS parameters that map distances to pixels. Thus, changing the appearance and paragraph layout of text~\cite{beier2022readability}.
\textbf{Character spacing} is the the distance between adjacent characters~\cite{cai2022personalized,suhail2026personalizing}. \textbf{Word spacing} is the space between adjacent words~\cite{beier2022readability,wallace2022space}.  \textbf{Line spacing} is the the distance between successive lines of text~\cite{rello2016make}. %

\textbf{Character opacity} assigns a unique opacity value to each rendered character~\cite{gu2024ai} according to

\begin{equation}
\alpha_i = 1 - u_i r
\label{eq:opacity_jitter}
\end{equation}

where $\alpha_i$ is the opacity of character $i$, $r \in [0,1]$ is the opacity jitter parameter, and $u_i$ is sampled uniformly from $[0,1]$. Increasing $r$ increases variation in character opacity. 

\subsubsection{Image-Level Perturbations}

Following text rendering, AIRRI applies configurable image-level perturbations directly to the rendered canvas to further reduce machine readability. AIRRI implements two image-level perturbation techniques: stripe overlays and image noise.

\textbf{Stripe Overlays} create periodic linear patterns across the rendered image. AIRRI generates a repeating stripe pattern with fixed stripe and gap widths using a configurable orientation angle $\theta$ and opacity parameter $\alpha_s$. AIRRI rotates the stripe pattern by angle $\theta$ and composites it onto the rendered image.

For \textbf{Image Noise}, AIRRI applies noise by superimposing randomly positioned circular artifacts to the rendered image. A user-defined parameter $\alpha_n \in [0,1]$ controls both artifact opacity and noise density. Given an image of width $W$ and height $H$, the number of generated artifacts is

\begin{equation}
N = \left\lfloor \frac{WH}{625}\alpha_n \right\rfloor
\label{eq:image_noise}
\end{equation}

where $N$ is the total number of noise artifacts. Increasing $\alpha_n$ increases both artifact density and visibility, resulting in greater disruption of automated text extraction \cite{noise}.

\subsubsection{Adversarial Perturbations}

We evaluate three black-box adversarial attacks against image-to-text systems. Each attack operates under a hard-label threat model: the attacker observes only the final discrete output of the image-to-text engine (i.e., the recognized text string), not internal confidence scores or gradients.
  
\textbf{RayS: Ray Searching Attack} \cite{chen2020rays} is a hard-label $\ell_\infty$ attack that searches for adversarial examples along ray directions in the input space. It maintains a discrete sign vector $\mathbf{s}_t \in \{-1,    +1\}^d$ and a boundary distance $d_t$. At iteration $t$, a candidate sign vector $\hat{\mathbf{s}}$ is generated by flipping signs within hierarchically determined blocks. The    adversarial candidate is    constructed along the corresponding ray:

\begin{equation}
\mathbf{x}_{\mathrm{adv}} = \mathrm{clip}\left(\mathbf{x} + d \cdot         \frac{\hat{\mathbf{s}}}{\|\hat{\mathbf{s}}\|},\; 0,\; 1\right)
\label{eq:rays_adv}
\end{equation}
  
A binary search determines the minimal distance $d^*$ at which the classifier decision $F$ changes:
            
\begin{equation}
d^* = \min \left\{ d \geq 0 \mid F\left(\mathrm{clip}\left(\mathbf{x} + d \cdot         \frac{\hat{\mathbf{s}}}{\|\hat{\mathbf{s}}\|},\; 0,\; 1\right)\right) \neq y        \right\}
\label{eq:rays_boundary}
\end{equation}
            
If $d^* < d_t$, the direction and distance are updated ($\mathbf{s}_{t+1} \leftarrow \hat{\mathbf{s}}$, $d_{t+1} \leftarrow d^*$). When this formulation finds a valid perturbation in the epsilon budget, it halts and bypasses gradient estimation.


\textbf{SurFree} \cite{maho2021surfree} is a geometrical decision-based $\ell_2$ attack that minimizes perturbation distance by exploring boundaries along orthogonal circular paths. Given a valid adversarial point $\mathbf{x}_{\mathrm{adv}}$,        candidate perturbations are explored on a 2D subspace defined by the unit        direction to the current adversarial point $\mathbf{d}_1$ and a random       orthogonal direction $\mathbf{d}_2$:
            
\begin{equation}
\mathbf{x}_{\theta} = \mathbf{x} + r \cos\theta \cdot \mathbf{d}_1 + r          \sin\theta \cdot \mathbf{d}_2
\label{eq:surfree_circular}
\end{equation}
            
where $r = \|\mathbf{x}_{\mathrm{adv}} - \mathbf{x}\|_2$, and $\mathbf{d}_2$ is sampled from low-frequency DCT components and orthogonalized via Gram-Schmidt. The maximum search angle $\theta_{\max}$ is dynamically adjusted based on step success using a factor $\rho$:
            
\begin{equation}
\theta_{\max}^{(t+1)} = \begin{cases} \theta_{\max}^{(t)} / \rho & \text{if step            successful,} \\ \theta_{\max}^{(t)} \cdot \rho & \text{otherwise.} \end{cases}
\label{eq:surfree_theta}
\end{equation}
            
\textbf{ADBA} \cite{chen2024adba} is a query-efficient decision-based attack that avoids estimating exact decision boundaries. Rather than computing the boundary distance $g(\mathbf{d})$ for candidate directions, it compares directions with an Approximated Decision Boundary (ADB).  To compare two perturbation directions $\mathbf{d}_1$ and $\mathbf{d}_2$, ADBA queries the target model at a scale defined by the ADB. If $\mathbf{d}_2$ successfully changes the label while $\mathbf{d}_1$ fails, then:

\begin{equation}
g(\mathbf{d}_2) \leq \text{ADB} < g(\mathbf{d}_1)
\label{eq:adba_inequality}
\end{equation}
            
Thus, $\mathbf{d}_2$ requires a smaller perturbation and is selected. ADBA continuously optimizes the perturbation size, which can exhaust the query budget without early stopping.

\subsection{Evaluation Pipeline}
The AIRRI evaluation pipeline measures automated text extraction performance by comparing extracted text against ground-truth source material. The evaluation process has two stages: automated extraction and extraction scoring.


\subsubsection{Automated Extraction}
AIRRI evaluates machine resistance using three image-to-text systems: Tesseract, EasyOCR, and GOT-OCR2. We selected these systems to represent both traditional image-to-text architectures and modern transformer-based vision-language models.
For each rendered image, all three extraction systems generate an independent text transcription. We evaluate Tesseract using the English language model, page segmentation mode 6, and OCR engine mode 3. We evaluate EasyOCR using its English language model with paragraph-level text aggregation enabled. GOT-OCR2 is evaluated using the publicly available \texttt{stepfun-ai/GOT-OCR-2.0-hf} checkpoint.
        
\subsubsection{Extraction Scoring}
        
AIRRI compares the extracted text from each image-to-text system against its ground-truth assignment text provided in the dataset manifest. AIRRI evaluates extraction performance using Character Error Rate (CER) and Word Error Rate (WER), two standard text-recognition metrics. For character-level evaluation, we normalize both the extracted and ground-truth text by converting all characters to lowercase and removing whitespace. Character Error Rate (CER) is computed according to Equation~\ref{eq:character_error_rate}:

\begin{equation}
CER =
\frac{D_L(T_{gt},T_{pred})}
{|T_{gt}|}
\label{eq:character_error_rate}
\end{equation}
        
where $D_L(\cdot)$ denotes the Levenshtein edit distance \cite{levenshtein}, $T_{gt}$ is the ground-truth text, $T_{pred}$ is the extracted text, and $|T_{gt}|$ is the number of characters in the ground-truth text.
        
For word-level evaluation, we convert both texts to lowercase and tokenize them into whitespace-delimited word sequences. Word Error Rate (WER) is computed according to Equation~\ref{eq:word_error_rate}:
        
\begin{equation}
WER =
\frac{D_L(W_{gt},W_{pred})}
{|W_{gt}|}
\label{eq:word_error_rate}
\end{equation}
        
where $W_{gt}$ and $W_{pred}$ denote the ground-truth and extracted word sequences, respectively, and $|W_{gt}|$ is the total number of words in the reference text. Equations~\ref{eq:character_error_rate} and \ref{eq:word_error_rate} show that both metrics measure normalized edit distance between the extracted and reference text. Lower CER and WER values indicate more successful text extraction, while higher values indicate stronger resistance to automated extraction. AIRRI reports both metrics for each image-to-text system and perturbation configuration.

\subsubsection{LLM Performance Evaluation}
While CER and WER quantify extraction quality, AIRRI seeks to disrupt downstream assignment completion using LLMs. To evaluate this effect, we supply an LLM with text extracted by each image-to-text system and use it to complete the assignment. We evaluated downstream LLM performance using ChatGPT through the ChatGPT web interface in Instant mode in July 2026. We conducted each evaluation in a new Temporary Chat to prevent prior conversation context from influencing the response. The raw OCR output was provided without manual correction or preprocessing using the prompt, ``Complete the following assignment to the best of your ability using only the information provided. Provide only your final answer, without additional commentary or explanation.'' A single inference was performed for each OCR extraction. Because the ChatGPT interface was used rather than the OpenAI API, inference parameters such as temperature, top-$p$, and random seed were not user-configurable and therefore were not explicitly controlled.

We score responses with a three-point scale: 0 (incorrect), 1 (partially correct), and 2 (correct). To aggregate results across assignments, AIRRI computes the \textit{LLM Performance Score (LPS)}:
\begin{equation}
LPS = \frac{\sum_{i=1}^{N} s_i}{2N}
\label{eq:lps}
\end{equation}
where $s_i \in \{0,1,2\}$ is the score assigned to response $i$ and $N$ is the number of evaluated responses. LPS ranges from 0 to 1; higher values indicate more successful assignment completion. For consistency with CER and WER, AIRRI also reports the \textit{LLM Error Rate (LER)}, where higher LER values indicate greater resistance to downstream assignment completion:
\begin{equation}
LER = 1 - LPS
\label{eq:ler}
\end{equation}

\subsection{Dataset for Evaluation}
Our new evaluation dataset has nine higher-education (HE) assignments collected from instructors at two universities. Five assignments were from a University of Connecticut English instructor and two from a University of Rhode Island History instructor. One English assignment has three essay topics with distinct assignment texts; thus, we created three separate evaluation instances. The final dataset contains seven English assignments and two History assignments. For evaluation, each assignment was rendered as clean and AIRRI-protected images, and CER and WER were computed. Table~\ref{tab:dataset_stats} summarizes the introduced dataset. Evaluation was also performed on open-source datasets of 8th and 12th grade-level reading passages (8-and-12) provided by The Readability Consortium~\cite{wallace2022towards}.

\begin{table}[t]
\centering
\caption{Summary statistics for the educational assignment dataset for AIRRI's evaluation. It has nine instructor-provided assignments from English and History courses with varying document lengths.}
\begin{tabular}{lr}
\toprule
Statistic & Value \\
\midrule
Assignments & 9 \\
English Assignments & 7 \\
History Assignments & 2 \\
Total Words & 4,364 \\
Total Characters & 25,282 \\
Average Words & 485 \\
Median Words & 382 \\
Minimum Words & 220 \\
Maximum Words & 1,122 \\
\bottomrule
\end{tabular}
\label{tab:dataset_stats}
\end{table}

\section{Results}
We evaluate AIRRI's effectiveness in degrading automated text extraction. We first compare extraction performance on clean and AIRRI-protected documents, then analyze the effects of individual adversarial attacks and extraction systems using CER and WER. Unless otherwise specified, the AIRRI pipeline generated AIRRI-protected documents, including all typographic and image-level perturbations applied simultaneously.

\subsection{Typographic + Image-Level Perturbations}
Table~\ref{tab:image_results} summarizes extraction performance for clean and AIRRI-protected documents across all image-to-text systems on both the HE Assignment and 8-and-12 datasets. Results are reported using Character Error Rate (CER) and Word Error Rate (WER), higher values indicate greater extraction difficulty.

AIRRI increased extraction error rates for EasyOCR and Tesseract. EasyOCR CER increased from 6.9\% to 55.9\%, while WER increased from 24.9\% to 96.7\%. Tesseract showed the largest degradation: CER increased from 2.3\% to 100\% and WER from 10.7\% to 100\%. This indicates complete extraction failure on all protected documents in the evaluation set.

GOT-OCR2 was less affected by the applied perturbations than some of the other models. CER increased from 30.7\% to 48.1\%, while WER decreased slightly from 63.3\% to 61.9\%. Examination of individual assignments showed considerable variation in performance: some protected documents produced substantially higher error rates than their clean counterparts, whereas others produced lower error rates.

On the 8-and-12 dataset, AIRRI also degraded extraction across all systems. EasyOCR CER increases from 7.3\% to 81.1\%. GOT-OCR2 showed the largest increase: CER rising from 0.8\% to 90.4\% and WER from 6.9\% to 91.5\%. Tesseract CER increased from 1.0\% to 38.8\%, less degradation than on the HE Assignment, though WER still reached 99.9\%.

\begin{table*}[t]
\centering
\caption{Average extraction error rates for clean and AIRRI-protected documents. CER is Character Error Rate, and WER is Word Error Rate.}
\label{tab:image_results}

\begin{tabular}{llcccccc}
\toprule
\multirow{2}{*}{Model} & \multirow{2}{*}{Metric} &
\multicolumn{3}{c}{HE Assignment} &
\multicolumn{3}{c}{8-and-12} \\
\cmidrule(lr){3-5} \cmidrule(lr){6-8}
 & & Clean & Prot. & $\Delta$ & Clean & Prot. & $\Delta$ \\
\midrule
EasyOCR
& CER (\%) & 6.94 & 55.95 & +49.01 & 7.34 & 81.10 & +73.76 \\
& WER (\%) & 24.86 & 96.67 & +71.81 & 24.84 & 96.96 & +72.12 \\
\midrule
Tesseract
& CER (\%) & 2.26 & 100.00 & +97.74 & 0.97 & 38.76 & +37.79 \\
& WER (\%) & 10.65 & 100.00 & +89.35 & 2.64 & 99.93 & +97.29 \\
\midrule
GOT-OCR2
& CER (\%) & 30.70 & 48.09 & +17.39 & 0.83 & 90.42 & +89.59 \\
& WER (\%) & 63.26 & 61.90 & -1.36 & 6.91 & 91.54 & +84.63 \\
\bottomrule
\end{tabular}

\end{table*}

\subsubsection{Individual Image Transformation Performance}
To evaluate the effectiveness of individual image transformations, each transformation was applied independently at five intensity levels, including a clean baseline (0\%) and perturbation strengths of 25\%, 50\%, 75\%, and 100\%. The 100\% level corresponded either to the maximum supported perturbation value or an empirically selected upper bound intended to avoid excessive visual degradation, whichever was lower. Although stronger perturbations generally led to higher OCR error rates, excessively strong transformations may also reduce human readability. We measured OCR performance at each intensity level and compared it with the clean baseline. Table~\ref{tab:individual_perturbations} reports the mean increase in image-to-text error rate across all assignments, perturbation intensity levels, and extraction systems.

\begin{table}[t]
\centering
\caption{Mean increase in image-to-text error rate (\%) relative to clean images, averaged across all assignments, perturbation intensity levels, and extraction systems.}
\label{tab:individual_perturbations}
\begin{tabular}{lcc}
\hline
Perturbation & CER $\Delta$ (\%) & WER $\Delta$ (\%) \\
\hline
Jitter  & 6.27  & 17.04 \\
Noise   & 3.96  & 17.17 \\
Stripes & \textbf{35.90} & \textbf{38.31} \\
\hline
\end{tabular}
\end{table}

Among the evaluated image transformations, stripe overlays produced the largest degradation in automated text extraction performance, increasing CER by an average of 35.9\% and WER by 38.3\%. In comparison, opacity jitter and image noise resulted in substantially smaller increases in error rates. Although performance varied across assignments and perturbation intensities, stripe overlays consistently produced the largest increases in both CER and WER. 

\subsection{AIRRI Degrades LLM Performance}
While CER and WER measure OCR extraction quality, the ultimate objective of AIRRI is to disrupt downstream LLM-assisted assignment completion. Table~\ref{tab:ler} presents the resulting LLM Performance Score (LPS) and LLM Error Rate (LER) for clean and AIRRI-protected documents. Across all image-to-text systems, AIRRI substantially increased downstream assignment completion failure rates. The aggregate LER increased from 7.4\% for clean documents to 55.6\% for protected documents, indicating a substantial reduction in the ability of LLMs to successfully complete assignments using extracted text.

\begin{table}[t]
\centering
\caption{LLM Performance Score (LPS) and LLM Error Rate (LER) for clean and AIRRI-protected documents. Higher LER values indicate stronger resistance to downstream assignment completion.}
\label{tab:ler}
\begin{tabular}{llcc}
\toprule
Protection & Model & LPS & LER \\
\midrule
Clean & EasyOCR & 94.4 & 5.6 \\
Clean & Tesseract & 94.4 & 5.6 \\
Clean & GOT-OCR2 & 88.9 & 11.1 \\
Clean & All & 92.6 & 7.4 \\
\midrule
Protected & EasyOCR & 77.8 & 22.2 \\
Protected & Tesseract & 0.0 & 100.0 \\
Protected & GOT-OCR2 & 55.6 & 44.4 \\
Protected & All & 44.4 & 55.6 \\
\bottomrule
\end{tabular}
\end{table}

Tesseract exhibited the largest increase in LER, rising from 5.6\% on clean documents to 100.0\% on protected documents, consistent with the complete extraction failure observed in the image-to-text evaluation. EasyOCR and GOT-OCR2 demonstrated greater resilience, with LER increasing from 5.6\% to 22.2\% and from 11.1\% to 44.4\%. Overall, AIRRI consistently increased failure rates across all evaluated extraction systems, with the strongest effect observed for Tesseract.

\subsection{Localized Degradation, Global Readability}

We evaluate three black-box adversarial attacks against three OCR engines where ADBA and Rays operate under an $L_\infty$ perturbation budget with $\varepsilon \in \{8/255, 16/255, 32/255, 64/255\}$, while SurFree uses an $L_2$ distance threshold.
Each attack targets only selected lines; the rest of the document remains unmodified.
We report CER and WER at two evaluation scopes: \textit{full\_composite}, where OCR runs on the entire stitched document, and \textit{target\_region}, where OCR runs only on the attacked line crops. Table~\ref{tab:adv-full-composite} shows the full-composite results.
Because only a small portion of the document is perturbed, CER and WER remain low across all conditions.
GOT-OCR~\cite{gotocr} achieves near-zero CER at $\varepsilon \leq 32/255$, while EasyOCR and Tesseract are around 6--8\% CER regardless of the attack.
Even at the largest budget ($\varepsilon = 64/255$), full-composite CER does not exceed 8.44\% for any engine, showing the targeted perturbation strategy helps preserve some document readability.

\begin{table*}[t]
\centering
\caption{Full-composite CER/WER for adversarial perturbations.}
\label{tab:adv-full-composite}

\begin{tabular}{llccccc}
\toprule
\multirow{2}{*}{Engine} & \multirow{2}{*}{Attack} &
\multirow{2}{*}{$\varepsilon$} &
\multicolumn{2}{c}{HE Assignment} &
\multicolumn{2}{c}{8-and-12} \\
\cmidrule(lr){4-5} \cmidrule(lr){6-7}
 & & & CER (\%) & WER (\%) & CER (\%) & WER (\%) \\
\midrule
\multirow{8}{*}{EasyOCR}
 & \multirow{4}{*}{ADBA} & 8/255  & 6.38 & 28.54 & 5.48 & 27.05 \\
 &                       & 16/255 & 5.82 & 28.23 & 5.05 & 25.62 \\
 &                       & 32/255 & 5.65 & 28.33 & 4.80 & 23.49 \\
 &                       & 64/255 & 8.07 & 34.00 & 6.46 & 28.80 \\
\cmidrule(l){2-7}
 & \multirow{4}{*}{Rays} & 8/255  & 6.78 & 29.31 & 6.99 & 29.77 \\
 &                       & 16/255 & 6.87 & 29.48 & 7.10 & 30.06 \\
 &                       & 32/255 & 7.17 & 30.73 & 7.05 & 29.68 \\
 &                       & 64/255 & 8.44 & 35.59 & 7.89 & 32.81 \\
\midrule
\multirow{8}{*}{GOT-OCR}
 & \multirow{4}{*}{ADBA} & 8/255  & 0.00 & 1.02 & 0.00 & 0.00 \\
 &                       & 16/255 & 0.00 & 0.59 & 0.00 & 0.00 \\
 &                       & 32/255 & 0.00 & 0.73 & 0.00 & 0.00 \\
 &                       & 64/255 & 1.69 & 8.58 & 0.00 & 1.44 \\
\cmidrule(l){2-7}
 & \multirow{4}{*}{Rays} & 8/255  & 0.31 & 2.10 & 0.49 & 2.78 \\
 &                       & 16/255 & 0.41 & 2.32 & 0.61 & 3.18 \\
 &                       & 32/255 & 0.73 & 4.05 & 0.55 & 2.66 \\
 &                       & 64/255 & 2.09 & 10.79 & 1.45 & 7.00 \\
\midrule
\multirow{8}{*}{Tesseract}
 & \multirow{4}{*}{ADBA} & 8/255  & 6.47 & 28.47 & 4.83 & 28.44 \\
 &                       & 16/255 & 6.02 & 27.82 & 4.40 & 27.05 \\
 &                       & 32/255 & 5.72 & 27.02 & 4.15 & 24.95 \\
 &                       & 64/255 & 7.92 & 32.83 & 5.82 & 30.16 \\
\cmidrule(l){2-7}
 & \multirow{4}{*}{Rays} & 8/255  & 6.82 & 29.05 & 5.91 & 29.54 \\
 &                       & 16/255 & 6.98 & 29.12 & 6.03 & 30.61 \\
 &                       & 32/255 & 7.18 & 30.28 & 5.97 & 29.18 \\
 &                       & 64/255 & 8.26 & 34.51 & 8.00 & 36.65 \\
\bottomrule
\end{tabular}

\end{table*}

Table~\ref{tab:adv-target-region} isolates the attacked lines.
Here, the picture changes sharply.
On the HE Assignment dataset, ADBA at $\varepsilon = 64/255$ raises CER to 49.83\% (EasyOCR), 57.93\% (GOT-OCR), and 48.55\% (Tesseract), with WER exceeding 80\% across all three engines.
The 8-and-12 dataset shows a similar pattern: ADBA at $\varepsilon = 64/255$ raises CER to 44.81\% (EasyOCR), 53.40\% (GOT-OCR), and 43.02\% (Tesseract), with WER reaching 76.14\%, 78.11\%, and 70.79\%, respectively.
Rays also degrades target-region accuracy as $\varepsilon$ grows, but its effect is weaker: at $\varepsilon = 64/255$ on the HE Assignment dataset, CER reaches only 11.44\% (EasyOCR) and 25.72\% (GOT-OCR), with comparable values of 9.28\% and 23.40\% on the 8-and-12 dataset.
At lower budgets ($\varepsilon \leq 16/255$), both attacks produce minimal changes for EasyOCR and Tesseract on both datasets.

\begin{table*}[t]
\centering
\caption{Target-region CER/WER for adversarial perturbations.}
\label{tab:adv-target-region}

\begin{tabular}{llccccc}
\toprule
\multirow{2}{*}{Engine} & \multirow{2}{*}{Attack} &
\multirow{2}{*}{$\varepsilon$} &
\multicolumn{2}{c}{HE Assignment} &
\multicolumn{2}{c}{8-and-12} \\
\cmidrule(lr){4-5} \cmidrule(lr){6-7}
 & & & CER (\%) & WER (\%) & CER (\%) & WER (\%) \\
\midrule
\multirow{8}{*}{EasyOCR}
 & \multirow{4}{*}{ADBA} & 8/255  & 4.36 & 23.96 & 3.95 & 21.61 \\
 &                       & 16/255 & 4.34 & 23.99 & 3.81 & 20.81 \\
 &                       & 32/255 & 6.31 & 29.87 & 4.85 & 24.23 \\
 &                       & 64/255 & 49.83 & 80.07 & 44.81 & 76.14 \\
\cmidrule(l){2-7}
 & \multirow{4}{*}{Rays} & 8/255  & 4.58 & 25.15 & 4.42 & 23.29 \\
 &                       & 16/255 & 4.93 & 26.22 & 4.59 & 23.36 \\
 &                       & 32/255 & 5.83 & 29.31 & 5.34 & 25.30 \\
 &                       & 64/255 & 11.44 & 45.24 & 9.28 & 39.32 \\
\midrule
\multirow{8}{*}{GOT-OCR}
 & \multirow{4}{*}{ADBA} & 8/255  & 19.79 & 24.70 & 18.91 & 28.10 \\
 &                       & 16/255 & 19.77 & 24.73 & 18.79 & 27.36 \\
 &                       & 32/255 & 21.42 & 30.55 & 19.67 & 30.49 \\
 &                       & 64/255 & 57.93 & 80.26 & 53.40 & 78.11 \\
\cmidrule(l){2-7}
 & \multirow{4}{*}{Rays} & 8/255  & 19.97 & 25.88 & 19.31 & 29.63 \\
 &                       & 16/255 & 20.26 & 26.94 & 19.45 & 29.70 \\
 &                       & 32/255 & 21.02 & 29.99 & 20.08 & 31.48 \\
 &                       & 64/255 & 25.72 & 45.77 & 23.40 & 44.34 \\
\midrule
\multirow{8}{*}{Tesseract}
 & \multirow{4}{*}{ADBA} & 8/255  & 4.65 & 25.08 & 6.19 & 26.30 \\
 &                       & 16/255 & 4.40 & 23.47 & 6.06 & 25.55 \\
 &                       & 32/255 & 6.71 & 31.92 & 7.07 & 28.76 \\
 &                       & 64/255 & 48.55 & 80.29 & 43.02 & 70.79 \\
\cmidrule(l){2-7}
 & \multirow{4}{*}{Rays} & 8/255  & 4.99 & 25.80 & 4.44 & 23.80 \\
 &                       & 16/255 & 5.27 & 26.91 & 4.57 & 23.40 \\
 &                       & 32/255 & 5.95 & 28.82 & 5.37 & 23.99 \\
 &                       & 64/255 & 11.33 & 44.48 & 9.51 & 38.92 \\
\bottomrule
\end{tabular}

\end{table*}

\subsection{Interviews with Human Readability Experts}

We conducted semi-structured interviews with four domain experts in typography, design, and cognitive psychology. We showed each human readability expert our machine-resistant designs and their perturbation levels. The interviews focused on the designs' potential effects on reading speed and comprehension across different reading modes, such as interlude reading or skimming. The final part of the interviews covered how the different designs affect neurodivergent readers and future design ideas we should explore. We performed thematic coding to analyze qualitative results from the semi-structured interviews. This effort revealed three codes for creating human-readable perturbations of machine-resistant text: preserving letterform structure, line tracking, and adequate foreground-background contrast. 

Notably, all experts found the vertical stripes readable. This shows that our most machine-resistant stripe design was the most human-readable, in our experts' view. Overall, this highlights a primary theme for future work: finding designs that minimally hurt human perception while maximizing degradation of machine perception in image-to-text. For example, all experts found the diagonal stripes unreadable, especially for skimming. P1 remarked they ``break up the letterforms''. Thus, this would not be a recommended human-readable design. All experts also found the 50\% opacity jitter acceptable, with P2 likening it to a ``printer running out ink''.

Future work ideas included experimenting with typefaces that affect vertical text orientation (i.e., mono-type fonts), text over complex backgrounds, varying font types within words (P3 ``ransom notes''), and applying opacity jitter only to characters within a word. Notably, all experts urged future work to explore new design ideas as AI technologies evolve and to bridge gaps in optimizing machine resistance while minimally degrading the human readability of AIRRI's designs.

\section{Discussion}

\subsection{Adversarial Perturbation Analysis}
ADBA was the strongest attack in our experiments. At $\varepsilon = 64/255$, it pushed target-region WER above 70\% on both datasets and across all three engines. RayS also degraded recognition with $\varepsilon$, but its peak target-region CER fell short of ADBA's. On the HE Assignment dataset, Rays reached 25.72\% CER on GOT-OCR and 11.44\% on EasyOCR; on the 8-and-12 dataset, the values were 23.40\% and 9.28\%. This gap likely stems from their search strategies: ADBA uses adaptive boundary estimation with offspring sampling to concentrate perturbation energy near the decision boundary~\cite{chen2024adba}, while Rays relies on random directional search that converges more slowly when the perturbation space is large \cite{chen2020rays}. This performance gap aligns with the findings in the original ADBA study~\cite{chen2024adba}, where ADBA consistently outperformed RayS on ImageNet. Our results indicate that this comparative advantage extends to document images and OCR models, and holds across both datasets. SurFree, the only $L_2$-bounded attack we tested, failed to produce usable perturbations because the $L_2$ distance between the original and perturbed images exceeded the allowed threshold at all budgets, and is thus omitted from the tables.

A notable pattern in Table~\ref{tab:adv-target-region} is GOT-OCR's high CER on target-region crops at low perturbation budgets (e.g., $\varepsilon = 8/255$): 19.79\% on the HE Assignment dataset and 18.91\% on the 8-and-12 dataset, compared to roughly 4--6\% for EasyOCR and Tesseract on both datasets. GOT-OCR is a vision-language model designed to process full document pages ~\cite{gotocr}, and it appears to struggle when given small, isolated line crops that lack the surrounding page context it was trained on. Despite this weaker performance on isolated crops, GOT-OCR's full-composite CER remains near zero on both datasets, as shown in Table~\ref{tab:adv-full-composite}, which indicates the model leverages document-level context to compensate for localized noise. However, this contextual advantage disappears when the evaluation scope is narrowed to the attacked region alone.

\subsection{Image-Level Perturbation Analysis}
Across the evaluated dataset, AIRRI consistently increased extraction error rates for EasyOCR and Tesseract, with Tesseract exhibiting complete failure to extract on all protected documents. These findings suggest that the typographic and image-level perturbations employed by AIRRI effectively disrupt the assumptions underlying many traditional image-to-text pipelines~\cite{wang2011end, ocr}.

A notable observation is the difference between traditional image-to-text systems and the transformer-based GOT-OCR2 model~\cite{gotocr, afkari2025comprehensive}. While AIRRI substantially degraded the performance of EasyOCR and Tesseract, GOT-OCR2 exhibited greater resistance and more variable behavior. Some assignments experienced increased extraction errors, while others showed improved performance relative to the clean baseline. Notably, its aggregate WER decreased slightly from 63.26\% to 61.90\%, despite CER increasing substantially from 30.70\% to 48.09\%. This suggests that AIRRI affects character- and word-level errors differently for GOT-OCR2. One possible explanation is that the perturbations alter segmentation or contextual decoding in ways that reduce some word-level errors while increasing character-level corruption; however, the underlying mechanism was not isolated in our experiments.  

\subsection{Individual Image-Level Perturbation Analysis}
The individual perturbation experiments provide insight into which image-level transformations contribute most strongly to machine resistance. Among the evaluated techniques, stripe overlays consistently produced the largest increases in both CER and WER, substantially outperforming image noise and character opacity variation. These results suggest that structured visual modifications disrupt automated text extraction more effectively than localized pixel-level distortions~\cite{noise, raisi2020text}.

Character opacity variation exhibited a different pattern~\cite{gu2024ai}. Across low and moderate intensity levels, opacity jitter produced relatively little degradation in extraction performance. We observed substantial increases in extraction error only at the highest tested intensity level, suggesting a threshold effect. However, the highest intensity setting also noticeably reduced human readability, reflecting the tradeoff between visual accessibility and text modification documented in readability research\cite{watson2021improving, beier2022readability, rello2016make, text_accessibility, hosseinpour2026onereader}. As a result, while extreme opacity variation can disrupt automated extraction, it may be less practical than stripe overlays, which achieved substantial degradation without requiring equally aggressive visual distortion.

Although stripe overlays were the most effective individual transformation, the complete AIRRI pipeline had the greatest degradation in extraction performance. It combines all typographic and image-level perturbations. Thus, AIRRI produced higher error rates than any single transformation alone. This result suggests that different perturbations likely interfere with different stages of the extraction process, and that their combined effects provide stronger protection than isolated modifications.

\subsection{Impact on Downstream LLM Performance}
While CER and WER quantify extraction quality, the LLM evaluation shows that extraction errors do not always translate directly into downstream task failure. Tesseract exhibited complete extraction failure on protected documents, resulting in an LER of 100\%. EasyOCR and GOT-OCR2 frequently produced noisy or incomplete extractions that retained sufficient information to reconstruct partial assignments, resulting in lower downstream failure rates. Our findings suggest that extraction-quality metrics alone may not fully capture the practical impact of extraction errors on modern AI-assisted workflows. 

Future work could explore human-inspectable AI-assisted workflows~\cite{yi2026data}. Although LLMs can often infer missing information and compensate for extraction mistakes, AIRRI increased the aggregate LER from 7.4\% to 55.6\%, indicating a substantial reduction in successful assignment completion. This demonstrates that sufficiently degraded extractions can still prevent reconstruction of assignment content, suggesting that AIRRI remains effective even against downstream AI-assisted recovery.

\subsection{Image-Level vs. Adversarial Perturbations}
This section compares the effectiveness and computational characteristics of AIRRI's image-level and adversarial perturbations. While both approaches aim to degrade automated text extraction, image-level perturbations apply across the entire rendered document, whereas adversarial perturbations introduce bounded pixel-level modifications within selected text regions.

\subsubsection{Effectiveness Against Image-to-Text Systems}
Overall, our results suggest that image-level perturbations provide the strongest protection in our setting. Among the evaluated adversarial methods, ADBA achieved the strongest extraction degradation, particularly at $\varepsilon = 64/255$. However, the image-level perturbations generally produced larger and more consistent increases in extraction error across the evaluated image-to-text systems. A notable exception was GOT-OCR2, where ADBA produced higher target-region error rates than those observed under the image-level perturbations. ADBA's strong performance against GOT-OCR2 suggests that adversarial approaches may offer advantages against more complex image-to-text systems, although additional evaluation of modern vision-language models is needed to confirm this.

\subsubsection{Hardware and Deployment Considerations}
Image-level perturbations are computationally inexpensive because they are applied directly during rendering. Let $n = WH$ denote the number of image pixels. Operations such as spacing adjustments, stripe overlays, opacity variation, and image noise require only a single pass over the image and therefore have time complexity $O(n)$.

In contrast, decision-based adversarial attacks require repeated queries to the target OCR system. Let $q$ denote the number of queries. Methods such as RayS~\cite{chen2020rays} and ADBA~\cite{chen2024adba} therefore scale as approximately $O(qn)$, since each query evaluates a perturbed image containing $n$ pixels. In our experiments, adversarial attacks often required thousands of queries, making them substantially more expensive than image-level perturbations.

Consequently, image-level and typographic perturbations provide the most practical and scalable approach for educational deployment. This distinction is particularly important for educational settings, where instructors may need to generate protected materials frequently and with minimal computational overhead. For this use case, it is impractical. Because adversarial attacks require substantially greater computation and generation time~\cite{brendel2018, chen2020hopskip, chen2020rays, chen2024adba}, they may be better suited for protecting selected high-value content rather than routine assignment generation.

\section{Conclusion}
We present AIRRI, a framework for generating machine-resistant educational content. AIRRI combines typographic perturbations, image-level perturbations, and decision-based adversarial attacks to reduce the effectiveness of automated text extraction systems while preserving access to the underlying instructional material. We evaluated AIRRI using a dataset of higher-education assignments. AIRRI substantially degraded the extraction performance for EasyOCR and Tesseract and produced more variable effects for the transformer-based GOT-OCR2 model. AIRRI also increased downstream assignment completion failure rates; thus, degraded extractions can help reduce the effectiveness of image-to-text-to-LLM workflows. Future work will evaluate AIRRI using formal human-readability studies~\cite{beier2022readability,tensmeyer2023web}, in-the-wild studies with educators and designers~\cite{wallace2021case,wallace2026attract}, larger educational datasets, datasets across domains like medical texts~\cite{ball2021optimizing}, and additional vision-language models to better understand the tradeoff between human usability and machine resistance.

\appendix

\section{Perturbation Level Parameters}
\label{app:perturbation_levels}

Table~\ref{tab:perturbation_levels} reports the parameter values used for
the four non-zero intensity levels of each AIRRI image-level perturbation.
These levels correspond to the perturbation examples and experimental
conditions evaluated in Figure~\ref{fig:image_perturbation_levels}.

\begin{table}[h]
\centering
\caption{Parameter values corresponding to AIRRI perturbation levels.
Level 0 represents the clean, unperturbed baseline.}
\label{tab:perturbation_levels}
\begin{tabular}{lccccc}
\toprule
Perturbation  & Level 1 & Level 2 & Level 3 & Level 4 \\
\midrule
Opacity Jitter  & 25 & 50 & 75 & 100 \\
Image Noise     & 25 & 50 & 75 & 100 \\
Stripe Overlay  & 10 & 20 & 40 & 60 \\
\bottomrule
\end{tabular}
\end{table}

\section{Full Pipeline Configuration}
\label{app:pipeline_configuration}

Table~\ref{tab:pipeline_configuration} reports the AIRRI configuration used
for the full-pipeline evaluation. These parameters were applied simultaneously
to generate the AIRRI-protected documents evaluated against the image-to-text
systems.

\begin{table}[h]
\centering
\caption{AIRRI configuration used for the full-pipeline evaluation.}
\label{tab:pipeline_configuration}
\begin{tabular}{lc}
\toprule
Parameter & Value \\
\midrule
Font Size & 16 px \\
Character Spacing & 3 px \\
Word Spacing & 15 px \\
Line Spacing & Default \\
Stripe Intensity & 20\% \\
Stripe Angle & $25^\circ$ \\
Opacity Jitter & 0.6 \\
Image Noise & 0 \\
\bottomrule
\end{tabular}
\end{table}

\bibliographystyle{plain}
\bibliography{main}

@inproceedings{gpt3,
 author = {Brown, Tom and Mann, Benjamin and Ryder, Nick and Subbiah, Melanie and Kaplan, Jared D and Dhariwal, Prafulla and Neelakantan, Arvind and Shyam, Pranav and Sastry, Girish and Askell, Amanda and Agarwal, Sandhini and Herbert-Voss, Ariel and Krueger, Gretchen and Henighan, Tom and Child, Rewon and Ramesh, Aditya and Ziegler, Daniel and Wu, Jeffrey and Winter, Clemens and Hesse, Chris and Chen, Mark and Sigler, Eric and Litwin, Mateusz and Gray, Scott and Chess, Benjamin and Clark, Jack and Berner, Christopher and McCandlish, Sam and Radford, Alec and Sutskever, Ilya and Amodei, Dario},
 booktitle = {Advances in Neural Information Processing Systems},
 editor = {H. Larochelle and M. Ranzato and R. Hadsell and M.F. Balcan and H. Lin},
 pages = {1877--1901},
 publisher = {Curran Associates, Inc.},
 title = {Language Models are Few-Shot Learners},
 url = {https://proceedings.neurips.cc/paper_files/paper/2020/file/1457c0d6bfcb4967418bfb8ac142f64a-Paper.pdf},
 volume = {33},
 year = {2020}
}

@article{academic_integrity,
author = {Sullivan, Miriam and Kelly, Andrew and McLaughlan, Paul},
year = {2023},
month = {03},
pages = {},
title = {ChatGPT in higher education: Considerations for academic integrity and student learning},
volume = {6},
journal = {Journal of Applied Learning \& Teaching},
doi = {10.37074/jalt.2023.6.1.17}
}

@article{academic_integrity2,
title = {Use of ChatGPT in academia: Academic integrity hangs in the balance},
journal = {Technology in Society},
volume = {75},
pages = {102370},
year = {2023},
issn = {0160-791X},
doi = {https://doi.org/10.1016/j.techsoc.2023.102370},
url = {https://www.sciencedirect.com/science/article/pii/S0160791X23001756},
author = {Saeed Awadh Bin-Nashwan and Mouad Sadallah and Mohamed Bouteraa}
}

@article{genai_guidelines,
title = {Generative AI in higher education: Seeing ChatGPT through universities' policies, resources, and guidelines},
journal = {Computers and Education: Artificial Intelligence},
volume = {7},
pages = {100326},
year = {2024},
issn = {2666-920X},
doi = {https://doi.org/10.1016/j.caeai.2024.100326},
url = {https://www.sciencedirect.com/science/article/pii/S2666920X24001292},
author = {Hui Wang and Anh Dang and Zihao Wu and Son Mac}
}

@misc{detecting_ai,
      title={Can AI-Generated Text be Reliably Detected?}, 
      author={Vinu Sankar Sadasivan and Aounon Kumar and Sriram Balasubramanian and Wenxiao Wang and Soheil Feizi},
      year={2025},
      eprint={2303.11156},
      archivePrefix={arXiv},
      primaryClass={cs.CL},
      url={https://arxiv.org/abs/2303.11156}, 
}

@article{ocr,
author = {Chaudhuri, Arindam and Mandaviya, Krupa and Badelia, Pratixa and Ghosh, Soumya},
year = {2017},
month = {01},
pages = {},
title = {Optical Character Recognition Systems for Different Languages with Soft Computing},
volume = {352},
isbn = {978-3-319-50251-9},
journal = {Studies in Fuzziness and Soft Computing},
doi = {10.1007/978-3-319-50252-6}
}

@inproceedings{text_accessibility,
author = {Somai, Mehek S. and Peiris, Roshan L and Tigwell, Garreth W.},
title = {A Review of Text Accessibility Standards, Guidelines, and Font Tool Limitations},
year = {2025},
isbn = {9798400706769},
publisher = {Association for Computing Machinery},
address = {New York, NY, USA},
url = {https://doi.org/10.1145/3663547.3759692},
doi = {10.1145/3663547.3759692},
booktitle = {Proceedings of the 27th International ACM SIGACCESS Conference on Computers and Accessibility},
articleno = {117},
numpages = {7},
location = {
},
series = {ASSETS '25}
}

@inproceedings{chen2020rays,
   title     = {{RayS}: A Ray Searching Method for Hard-Label Adversarial Attack},
   author    = {Chen, Jinghui and Gu, Quanquan},
   booktitle = {Proceedings of the 26th ACM SIGKDD International Conference on Knowledge Discovery \& Data Mining (KDD)},
   year      = {2020},
 }

@inproceedings{maho2021surfree,
   title     = {{SurFree}: a fast surrogate-free black-box attack},
   author    = {Maho, Thibault and Furon, Teddy and Le Merrer, Erwan},
   booktitle = {Proceedings of the IEEE/CVF Conference on Computer Vision and Pattern Recognition (CVPR)},
   year      = {2021},
 }

@misc{chen2024adba,
   title     = {{ADBA}: Approximated Decision Boundary Approach for Black-box Adversarial Attacks},
   author    = {Chen, Feiyang and Wang, Dong and Xu, Xin and Gao, Jing},
   year      = {2024},
   eprint    = {2406.04998},
   archivePrefix={arXiv},
 }

@misc{noise,
      title={Quality of OCR for Degraded Text Images}, 
      author={Roger T. Hartley and Kathleen Crumpton},
      year={1999},
      eprint={cs/9902009},
      archivePrefix={arXiv},
      primaryClass={cs.DL},
      url={https://arxiv.org/abs/cs/9902009}, 
}

@INPROCEEDINGS{tesseract,
  author={Smith, R.},
  booktitle={Ninth International Conference on Document Analysis and Recognition (ICDAR 2007)}, 
  title={An Overview of the Tesseract OCR Engine}, 
  year={2007},
  volume={2},
  number={},
  pages={629-633},
  doi={10.1109/ICDAR.2007.4376991}}

@misc{gotocr,
      title={General OCR Theory: Towards OCR-2.0 via a Unified End-to-end Model}, 
      author={Haoran Wei and Chenglong Liu and Jinyue Chen and Jia Wang and Lingyu Kong and Yanming Xu and Zheng Ge and Liang Zhao and Jianjian Sun and Yuang Peng and Chunrui Han and Xiangyu Zhang},
      year={2024},
      eprint={2409.01704},
      archivePrefix={arXiv},
      primaryClass={cs.CV},
      url={https://arxiv.org/abs/2409.01704}, 
}

@misc{easyocr,
  author = {{Jaided AI}},
  title = {EasyOCR},
  year = {2020},
  howpublished = {\url{https://github.com/JaidedAI/EasyOCR}},
  note = {Accessed: 2026-06-13}
}

@article{levenshtein,
  author = {Levenshtein, Vladimir I.},
  title = {Binary Codes Capable of Correcting Deletions, Insertions, and Reversals},
  journal = {Soviet Physics Doklady},
  volume = {10},
  number = {8},
  pages = {707--710},
  year = {1966}
}

@ARTICLE{traditional_adversarial,
  author={Costa, Joana C. and Roxo, Tiago and Proença, Hugo and Inácio, Pedro Ricardo Morais},
  journal={IEEE Access}, 
  title={How Deep Learning Sees the World: A Survey on Adversarial Attacks \& Defenses}, 
  year={2024},
  volume={12},
  number={},
  pages={61113-61136},
  doi={10.1109/ACCESS.2024.3395118}
}

@article{information_retrieval,
author = {Vechtomova, Olga},
year = {2009},
month = {06},
pages = {307-309},
title = {Introduction to Information Retrieval Christopher D. Manning, Prabhakar Raghavan, and Hinrich Schütze (Stanford University, Yahoo! Research, and University of Stuttgart) Cambridge: Cambridge University Press, 2008, xxi+482 pp; hardbound, ISBN 978-0-521-86571-5, \$60.00},
volume = {35},
journal = {Computational Linguistics},
doi = {10.1162/coli.2009.35.2.307}
}

@inproceedings{cai2022personalized,
  title={Personalized font recommendations: Combining ML and typographic guidelines to optimize readability},
  author={Cai, Tianyuan and Wallace, Shaun and Rezvanian, Tina and Dobres, Jonathan and Kerr, Bernard and Berlow, Samuel and Huang, Jeff and Sawyer, Ben D and Bylinskii, Zoya},
  booktitle={Proceedings of the 2022 ACM designing interactive systems conference},
  pages={1--25},
  year={2022}
}

@inproceedings{suhail2026personalizing,
  title={Personalizing Font Size Normalization: Linguistic Background Shapes Which Font Attributes Matter},
  author={Suhail, Yusra and Khadka, Kushas and Llana, Borano and Hosseinpour, Maedeh and Wallace, Shaun},
  booktitle={Proceedings of the 34th ACM Conference on User Modeling, Adaptation and Personalization},
  pages={203--212},
  year={2026}
}

@article{wallace2021case,
	title        = {Case Studies on the Motivation and Performance of Contributors Who Verify and Maintain In-Flux Tabular Datasets},
	author       = {Wallace, Shaun and Papoutsaki, Alexandra and Tan, Neilly H and Guo, Hua and Huang, Jeff},
	year         = 2021,
	journal      = {Proceedings of the ACM on Human-Computer Interaction},
	publisher    = {ACM New York, NY, USA},
	volume       = 5,
	number       = {CSCW2},
	pages        = {1--25}
}

@article{wallace2026attract,
  title={Attract \& Engage Visitors for Tabular Data Maintenance: a Longitudinal Naturalistic Field Study},
  author={Wallace, Shaun and Lee, Na Kyoung and Peng, Zhengyi and Massachi, Talie and Do, Long and Rathore, Surbhi and Brown, Sarah M and Huang, Jeff},
  journal={Proceedings of the ACM on Human-Computer Interaction},
  volume={10},
  number={6},
  pages={1--41},
  year={2026},
  publisher={ACM New York, NY}
}

@article{wallace2022space,
  title={Space for readability: effects on reading speed from individuated character and word spacing},
  author={Wallace, Shaun and Dobres, Jonathan and Bylinskii, Zoya and Sawyer, Ben},
  journal={Journal of Vision},
  volume={22},
  number={14},
  pages={3349--3349},
  year={2022},
  publisher={The Association for Research in Vision and Ophthalmology}
}

@article{beier2022readability,
  title={Readability research: An interdisciplinary approach},
  author={Beier, Sofie and Berlow, Sam and Boucaud, Esat and Bylinskii, Zoya and Cai, Tianyuan and Cohn, Jenae and Crowley, Kathy and Day, Stephanie L and Dingler, Tilman and Dobres, Jonathan and others},
  journal={Foundations and Trends{\^W} in Accounting},
  volume={16},
  number={4},
  pages={214--324},
  year={2022},
  publisher={Emerald Publishing Limited}
}

@inproceedings{rello2016make,
  title={Make it big! The effect of font size and line spacing on online readability},
  author={Rello, Luz and Pielot, Martin and Marcos, Mari-Carmen},
  booktitle={Proceedings of the 2016 CHI conference on Human Factors in Computing Systems},
  pages={3637--3648},
  year={2016}
}

@inproceedings{gu2024ai,
  title={An AI-resilient text rendering technique for reading and skimming documents},
  author={Gu, Ziwei and Arawjo, Ian and Li, Kenneth and Kummerfeld, Jonathan K and Glassman, Elena L},
  booktitle={Proceedings of the 2024 CHI Conference on Human Factors in Computing Systems},
  pages={1--22},
  year={2024}
}

@article{raisi2020text,
  title={Text detection and recognition in the wild: A review},
  author={Raisi, Zobeir and Naiel, Mohamed A and Fieguth, Paul and Wardell, Steven and Zelek, John},
  journal={arXiv preprint arXiv:2006.04305},
  year={2020}
}

@inproceedings{wang2011end,
  title={End-to-end scene text recognition},
  author={Wang, Kai and Babenko, Boris and Belongie, Serge},
  booktitle={2011 International conference on computer vision},
  pages={1457--1464},
  year={2011},
  organization={IEEE}
}

@article{afkari2025comprehensive,
  title={A comprehensive survey of transformers in text recognition: Techniques, challenges, and future directions},
  author={Afkari-Fahandari, Ali and Shabaninia, Elham and Asadi-Zeydabadi, Fatemeh and Nezamabadi-Pour, Hossein},
  journal={ACM Computing Surveys},
  volume={58},
  number={5},
  pages={1--42},
  year={2025},
  publisher={ACM New York, NY}
}

@article{wallace2022towards,
  title={Towards individuated reading experiences: Different fonts increase reading speed for different individuals},
  author={Wallace, Shaun and Bylinskii, Zoya and Dobres, Jonathan and Kerr, Bernard and Berlow, Sam and Treitman, Rick and Kumawat, Nirmal and Arpin, Kathleen and Miller, Dave B and Huang, Jeff and others},
  journal={ACM Transactions on Computer-Human Interaction (TOCHI)},
  volume={29},
  number={4},
  pages={1--56},
  year={2022},
  publisher={ACM New York, NY}
}

@misc{kiss2019ocr,
      title={Brno Mobile OCR Dataset}, 
      author={Martin Kišš and Michal Hradiš and Oldřich Kodym},
      year={2019},
      eprint={1907.01307},
      archivePrefix={arXiv},
      primaryClass={cs.CV},
      url={https://arxiv.org/abs/1907.01307}, 
}

@misc{2021ocrCER,
  title = {3D Rendering Framework for Data Augmentation in Optical Character Recognition : (Invited Paper)},
  author = {Andreas Spruck and Maximiliane Hawesch and Anatol Maier and C. Riess and Jürgen Seiler and André Kaup},
  year = {2021},
  url = {https://arxiv.org/pdf/2209.14970},
  doi = {10.1109/ISSCS52333.2021.9497438}
}

@misc{brendel2018,
      title={Decision-Based Adversarial Attacks: Reliable Attacks Against Black-Box Machine Learning Models}, 
      author={Wieland Brendel and Jonas Rauber and Matthias Bethge},
      year={2018},
      eprint={1712.04248},
      archivePrefix={arXiv},
      primaryClass={stat.ML},
      url={https://arxiv.org/abs/1712.04248}, 
}

@misc{chen2020hopskip,
      title={HopSkipJumpAttack: A Query-Efficient Decision-Based Attack}, 
      author={Jianbo Chen and Michael I. Jordan and Martin J. Wainwright},
      year={2020},
      eprint={1904.02144},
      archivePrefix={arXiv},
      primaryClass={cs.LG},
      url={https://arxiv.org/abs/1904.02144}, 
}

@misc{googlefontsanalytics,
  author       = {{Google Fonts}},
  title        = {Google Fonts Analytics},
  howpublished = {\url{https://fonts.google.com/analytics}},
  note         = {Accessed: 2026-08-18}
}

@inproceedings{llana2026reading,
author = {Llana, Borano and Baron, Alisa and Yu, Haihan and Hosseinpour, Maedeh and Suhail, Yusra and Chin, Sean and Khadka, Kushas and Wallace, Shaun},
title = {Reading with Diversity in Mind: Pupillometry and Typography Towards Inclusive Design for ADHD Readers},
year = {2026},
isbn = {9798400722813},
publisher = {Association for Computing Machinery},
address = {New York, NY, USA},
url = {https://doi.org/10.1145/3772363.3799383},
doi = {10.1145/3772363.3799383},
booktitle = {Proceedings of the Extended Abstracts of the 2026 CHI Conference on Human Factors in Computing Systems},
articleno = {549},
numpages = {7},
location = {
},
series = {CHI EA '26}
}

@inproceedings{palmen2023bold,
author = {Palm{\'e}n, Hilary and Gilbert, Michael and Crossland, David},
title = {How bold can we be? The impact of adjusting font grade on readability in light and dark polarities},
year = {2023},
isbn = {9781450394215},
publisher = {Association for Computing Machinery},
address = {New York, NY, USA},
url = {https://doi.org/10.1145/3544548.3581552},
doi = {10.1145/3544548.3581552},
booktitle = {Proceedings of the 2023 CHI Conference on Human Factors in Computing Systems},
articleno = {402},
numpages = {11},
location = {Hamburg, Germany},
series = {CHI '23}
}

@inproceedings{niklaus2023digital,
  title={Digital reading rulers: Evaluating inclusively designed rulers for readers with dyslexia and without},
  author={Niklaus, Aleena Gertrudes and Cai, Tianyuan and Bylinskii, Zoya and Wallace, Shaun},
  booktitle={Proceedings of the 2023 CHI conference on human factors in computing systems},
  pages={1--17},
  year={2023}
}

@inproceedings{yi2026data,
  title={Data Canvas: A Provenance-Guided Harness for Agentic Data Engineering},
  author={Yi, Zixuan and Shao, Yuanming and Wallace, Shaun and Ives, Zachary and Marcus, Ryan},
  booktitle={Third Conference on Language Modeling},
  pages={1--13},
  year={2026}
}

@inproceedings{suhail2026fontattributes,
  author    = {Suhail, Yusra and Khadka, Kushas and Llana, Borano and Hosseinpour, Maedeh and Salaun, Emory and Wallace, Shaun},
  title     = {Different Font Attributes Influence English Readers' Perception of Font Size Across Age Groups},
  booktitle = {CHI 2026 Workshop on Science and Technology for Augmenting Reading (STAR)},
  year      = {2026},
  address   = {Barcelona, Spain},
  date      = {2026-04-16},
  url       = {https://chi-star-workshop.github.io/src/assets/pdf/papers/CHI_Workshop___Font_Size_Normalization_by_Age-5%20-%20Yusra%20Suhail.pdf},
  note      = {Workshop paper}
}

@article{llana2025typography,
  title={Typography and Pupilometry: Exploring the Speed, Comprehension, and Concentration Trade-Off in Reading},
  author={Llana, Borano and Baron, Alisa and Khadka, Kushas and Suhail, Yusra and Wallace, Shaun},
  journal={Journal of Vision},
  volume={25},
  number={9},
  pages={2470--2470},
  year={2025},
  publisher={The Association for Research in Vision and Ophthalmology}
}

@inproceedings{nguyen2026rears,
  author    = {Nguyen, ThaoAn Lily and Llana, Borano and Piccoli, Jonathan and Suhail, Yusra and Hosseinpour, Maedeh and Wallace, Shaun},
  title     = {REARS: An Interface for Browser-Based Eye-Tracking in Remote Readability Studies},
  booktitle = {CHI 2026 Workshop on Science and Technology for Augmenting Reading (STAR)},
  year      = {2026},
  address   = {Barcelona, Spain},
  date      = {2026-04-16},
  url       = {https://chi-star-workshop.github.io/src/assets/pdf/papers/CHI_STAR_REARS_FINAL%20-%20Borano%20Llana.pdf},
  note      = {Workshop paper}
}

@inproceedings{ball2021optimizing,
  title={Optimizing electronic health records through readability},
  author={Ball, Rachel V and Miller, Dave B and Wallace, Shaun and Macias, Kathlyn Camargo and Ibrahim, Mahmoud and Gonzaga, Ernesto Robalino and Karasik, Olga and Rohlsen-Neal, Dekai R and Barrientos, Sarah and Ross, Edward A and others},
  booktitle={Proceedings of the International Symposium on Human Factors and Ergonomics in Health Care},
  volume={10},
  number={1},
  pages={65--70},
  year={2021},
  organization={SAGE Publications Sage CA: Los Angeles, CA}
}

@inproceedings{hosseinpour2026onereader,
  author    = {Hosseinpour, Maedeh and Llana, Borano and Suhail, Yusra and Khadka, Kushas and Garcia, Grant and Wallace, Shaun},
  title     = {One Reader, Two Reading Profiles: Language-Dependent Dyslexia and Reading Behaviors in English-Spanish Readers},
  booktitle = {CHI 2026 Workshop on Science and Technology for Augmenting Reading (STAR)},
  year      = {2026},
  address   = {Barcelona, Spain},
  date      = {2026-04-16},
  url       = {https://chi-star-workshop.github.io/src/assets/pdf/papers/One%20Reader%20Two%20Reading%20Profiles_Hosseinpour%20-%20maedeh%20hp.pdf},
  note      = {Workshop paper}
}

@article{watson2021improving,
  title={Improving reading outcomes using digital reading rulers for readers with \& without dyslexia},
  author={Watson, Aleena and Wallace, Shaun},
  journal={Journal of Vision},
  volume={21},
  number={9},
  pages={2650--2650},
  year={2021},
  publisher={The Association for Research in Vision and Ophthalmology}
}

@inproceedings{tensmeyer2023web,
  title={Web Table Formatting Affects Readability on Mobile Devices},
  author={Tensmeyer, Christopher and Bylinski, Zoya and Cai, Tianyuan and Miller, Dave and Nenkova, Ani and Niklaus, Aleena and Wallace, Shaun},
  booktitle={Proceedings of the ACM Web Conference 2023},
  pages={1334--1344},
  year={2023}
}

\end{document}